\documentclass[letterpaper, 10 pt, conference]{IEEEtran}  

\IEEEoverridecommandlockouts                              

\usepackage{graphics} 
\usepackage{epsfig} 
\usepackage{mathptmx} 
\usepackage{times} 
\usepackage{amsmath} 
\usepackage{amssymb}  

\usepackage[T1]{fontenc}
\usepackage{amsfonts}
\usepackage{booktabs}
\usepackage{siunitx}
\usepackage{caption} 
\usepackage{booktabs}
\usepackage{graphicx}
\usepackage{algorithm, algpseudocode}
\usepackage{float}
\usepackage[percent]{overpic}
\usepackage{caption}
\DeclareGraphicsExtensions{.pdf,.png}

\usepackage[font=small,skip=0pt]{subcaption}

\usepackage{array, tabularx, makecell, booktabs}%

\usepackage{tikz}

\usepackage{geometry}
\usepackage{comment}

\title{\LARGE \bf
Holistic Deep-Reinforcement-Learning-based Training of Autonomous Navigation Systems
}

\author{Linh K{\"a}stner$^{1}$\thanks{$^{1}$Linh K{\"a}stner, Marvin Meusel, Teham Bhuiyan, and Jens Lambrecht are with the Chair Industry Grade Networks and Clouds, Faculty of Electrical Engineering, and Computer Science,				
		Berlin Institute of Technology, Berlin, Germany
		{\tt\small linhdoan@win.tu-berlin.de}}, Marvin Meusel$^{1}$, Teham Bhuiyan$^{1}$, and Jens Lambrecht$^{1}$
}

\begin{document}

\maketitle
\thispagestyle{empty}
\pagestyle{empty}


\begin{abstract}

In recent years, Deep Reinforcement Learning emerged as a promising approach for autonomous navigation of ground vehicles and has been utilized in various areas of navigation such as cruise control, lane changing, or obstacle avoidance. However, most research works either focus on providing an end-to-end solution training the whole system using Deep Reinforcement Learning or focus on one specific aspect such as local motion planning. This however, comes along with a number of problems such as catastrophic forgetfulness, inefficient navigation behavior, and non-optimal synchronization between different entities of the navigation stack. In this paper, we propose a holistic Deep Reinforcement Learning training approach in which the training procedure is involving all entities of the navigation stack. This should enhance the synchronization between- and understanding of all entities of the navigation stack and as a result, improve navigational performance. We trained several agents with a number of different observation spaces to study the impact of different input on the navigation behavior of the agent. In profound evaluations against multiple learning-based and classic model-based navigation approaches, our proposed agent could outperform the baselines in terms of efficiency and safety attaining shorter path lengths, less roundabout paths, and less collisions.

\end{abstract}
\section{Introduction}
\noindent As human-machine-collaboration becomes essential, mobile robot navigation in crowded environments is increasingly becoming an important aspect to consider. Traditional navigation stacks of ground vehicles used within industrial setups such as AGVs utilize the ROS navigation stack \cite{marder2010office}, which consists of a global planner, which, given a global map, calculates an optimal path from a start point to a goal position and a local planner, which executes the global plan by utilizing sensor observations to avoid dynamic obstacles that were not present in the map. While navigation in static or slightly dynamic environments can be solved with currently employed navigation stacks, navigation in highly dynamic environments remains a challenging task \cite{kastner-aio}. It requires the agent to efficiently generate safe actions in proximity to unpredictably moving obstacles in order to avoid collisions. Traditional model-based motion planning approaches often employ hand-engineered safety rules to avoid dynamic obstacles. However, hand-designing the navigation behavior in dense environments is difficult since the future motion of the obstacles is unpredictable \cite{qian2010socially}. 
\begin{figure}[!h]
    \centering
    \includegraphics[width=0.7\linewidth]{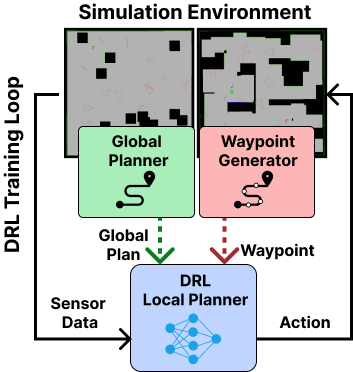}
    \caption{Information about the global planner and waypoint generator will be given as input for the DRL agent in order to enhance understanding of the DRL agent for high-level planners and thus improve synchronization and navigation efficiency.}
    \label{intro}
\end{figure}
\noindent In recent years, Deep Reinforcement Learning (DRL) has emerged as an end-to-end method that demonstrated superiority for obstacle avoidance in dynamic environments and for learning complex behavior rules. Thus, a variety research publications incorporated DRL to solve high-level tasks such as grasping, navigation or simulation \cite{chen2017socially}, \cite{kastner2021towards}, \cite{dugas2020navrep}, \cite{chen2019crowd}. However, DRL-based navigation approaches come along with issues such as difficult training, the myopic nature of the DRL agent, or catastrophic forgetfulness \cite{faust2018prm}, \cite{xiao2022motion}. Recent approaches either handled this problem by shortening the planning horizon using waypoints \cite{brito2021go},\cite{kastner2021connecting},  employing hybrid approaches \cite{faust2018prm} \cite{chiang2019learning}, or switch between classic model-based navigation and DRL planners. However, regarding the parts of the navigation system separately can lead to synchronization issues and non-optimal behavior such as jerky motions or the agent moving too far away from the initially planned global path. 
On that account, this paper proposes a holistic training approach incorporating the global planner and the waypoint generator into the DRL training pipeline. Therefore, classic global planners such as RRT or A*, and the waypoint generators presented in our previous work will be utlilized to provide the agent with more information about the higher-level planning directly within its training procedure.
Thus, the understanding of the agent for decisions made by other components of the navigation stack should be improved, which makes navigation smoother and more consistent. We compare different agent inputs and evaluate all agents against classic baseline approaches within the simulation platform arena-rosnav \cite{kastner2021towards} in terms of various navigational metrics. The main contributions of this work are the following:
\begin{itemize}
    \item Proposal of an holistic training approach utilizing the whole navigation stack instead of an isolated training procedure
    \item Incorporation of global planning and waypoint information into the reward system of the agent to improve synchronization between the entities and as a result improve navigational performance
    \item Qualitative and quantitative evaluation on different highly dynamic environments and comparison against a baseline DRL and classic model-based navigation approaches

\end{itemize}

\noindent The paper is structured as follows. Sec. II begins with related works. Subsequently, the methodology is presented in Sec. III. Sec. IV presents the results and discussion. Finally, Sec. V will provides a conclusion and outlook.

\section{Related Works}
\noindent DRL-based navigation approaches proved to be a promising alternative that has been successfully applied into various approaches for navigation of vehicles and robots with remarkable results. Various works demonstrated the superiority of DRL-based OA approaches due to more flexiblility in the handling of obstacles, generalization new problem instances, and ability to learn more complex tasks without manually designing the functionality. Thus, various research works incorporated DRL into their navigation systems for tasks such as lane changing \cite{ye2021meta}, cruise control \cite{yavas2022model}, \cite{chen2021automatic}, \cite{brosowsky2021safe}, cooperative behavior planning \cite{kurzer2022learning}, or and obstacle avoidance \cite{dugas2020navrep},  \cite{chen2019crowd}, \cite{kastner2021towards}. Atouri et al. \cite{atoui2022intelligent} proposed a DRL-based control switch for lateral control of autonomous vehicles. Similarily, Kästner et al. proposed a DRL-based control switch to choose between different navigation policies \cite{kastner-aio}. Liu et al. proposed a DRl approach for autonomous driving of vehicles in urban environments using expert demonstrations.
\\\\\noindent
Other works incorporated DRL for dynamic obstacle avoidance. Works from Everet et al. \cite{everett2018motion} and Chen et al. \cite{chen2019crowd} require the exact obstacle positions and perform a DRL-based obstacle avoidance approach. Dugas et al. relied solely on DRL for navigation \cite{dugas2020navrep}. The authors remarked that this could lead to jerky motions and failed navigation for long ranges. Since the reward that a DRL agent can obtain in long-range navigation over large-scale maps is usually sparse, agents are only suitable for short-range navigation due to local minima. Thus, a variety of research works combine DRL-based local planning with traditional methods such as RRT \cite{lavalle1998rapidly} or A-Star \cite{hart1968formal}. Faust et al. utilized DRL to assist an PRM-based global planner \cite{faust2018prm}. Similarily, Chiang et al. utilized DRL in combination with the RRT global planner. Other works utilize waypoints, as an interface for communication between global and local planner. These are points sampled from the global path to be given as input to the DRL agent in order to shorten its planning horizon. Gundelring et al. \cite{guldenringlearning} integrated a DRL-based local planner with a conventional global planner employing a simple subsampling of the global path given a static lookahead distance to create waypoints for the DRL-local planner. Similarly, Regler et al. \cite{regier2020deep} propose a hand-designed sub-sampling to deploy a DRL-based local planner with conventional navigation stacks.
A limitation of these works is that the simple sub-sampling of the global path is inflexible and could lead to hindrance in complex situations, e.g. when multiple humans are blocking the way. 
\\\\\noindent 
Hence, other works employed a more intelligent way to generate waypoints. Brito et al. \cite{brito2021go} proposed a DRL-based waypoint generation where the agent is trained to learn a cost-to-go model to directly generate subgoals, which an MPC planner follows. The better estimated cost-to-go value enables MPC to solve a long-term optimal trajectory. Similarly, Bansal et al. \cite{bansal2020combining} proposed a method called LB-WayPtNav, in which a supervised learning-based perception module is used to process RGB image data and output a waypoint. With the waypoint and robot current state, a spline-based smooth trajectory is generated and tracked by a traditional model-based, linear feedback controller to navigate to the waypoint. However, the proposed supervised training approach, requires a tedious data acquisition stage to provide annotated training data. In our previous work, we proposed various waypoint generation approaches that are more flexible \cite{kastner2021connecting}, \cite{kastner2021towards}, \cite{kastner2021obstacle} and could show improved navigation performance in long-range navigation within crowded environments.
\\\\\noindent
Of note, all previously mentioned works train the DRL agent as a separate entity and later incorporated it into the navigation stack, which could result in a number of issues such as synchronization problems and inefficient navigation behavior. The DRL agent is almost always trained for short-range obstacle avoidance and produce failures over long-ranges. Furthermore, navigation performance of the DRL agent is also heavily dependent on the efficiency of the global planner or the waypoint generator. On that account, this work incorporates all entities of the navigation stack into the training pipeline. More specifically, the whole navigation stack consisting of the global planner, the waypoint generator, and the DRL agent is deployed in the training pipeline. The DRL agent should still be responsible for local obstacle avoidance but receive high level input of the other two entities as input to improve its understanding of their decisions. Thus, a better synchronization and inter-operation between the three entities should be attained.

\begin{figure*}[]
    \centering
    \includegraphics[width=0.99\linewidth]{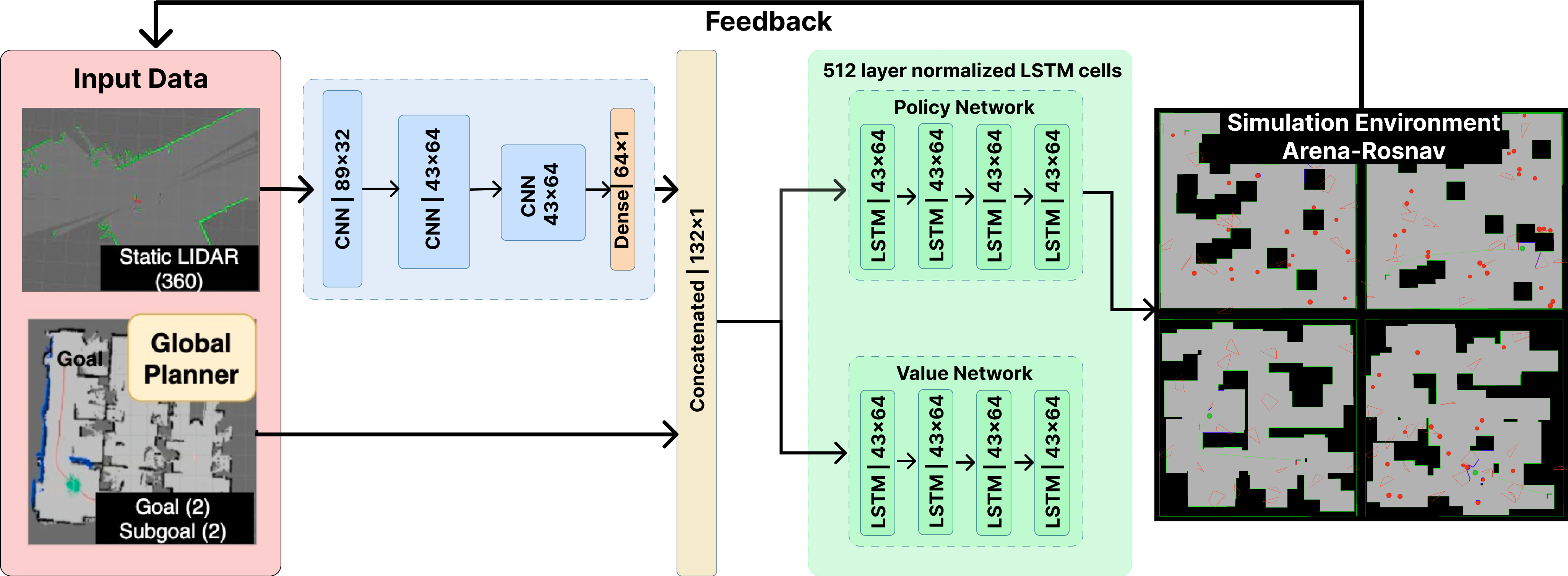}
    \caption{System Design and Training Pipeline. The input is exemplary for Agent 4. The specific parameters and tensor sizes for each of the agents are specified in Table \ref{tab:agent-input}}
    \label{system}
\end{figure*}

\section{Methodology}
\noindent In this chapter, we present the methodology of our proposed framework. In total six agents with different inputs are trained.

\subsection{System Design and training procedure}

\begin{figure*}[!h]
    \centering
    \includegraphics[width=0.99\linewidth]{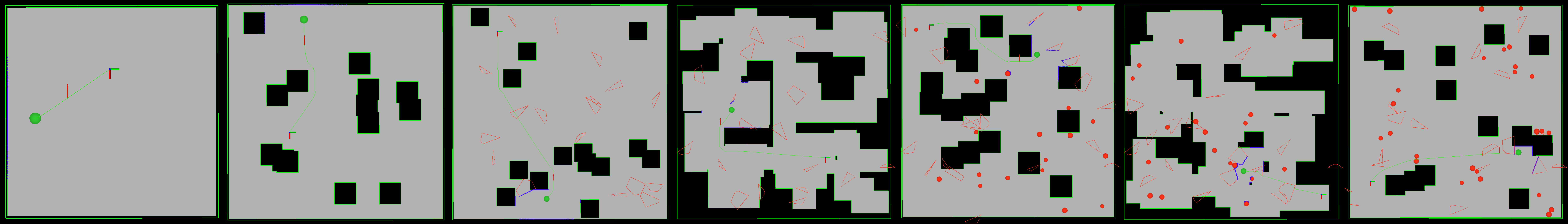}
    \caption{Stage one to seven.}
    \label{stages}
\end{figure*}

\noindent Fig. \ref{system} illustrates the system design of our approach. The DRL agents are trained within our 2D simulation environment arena-rosnav \cite{kastner2020deep} and trained using the staged training curriculum, that is whenever the agent reaches a succes rate of 80 percent the next stage with increasing difficulty will be loaded. The stages contain dynamic and static obstacles spawned randomly. The stages are illustrated in Fig. \ref{stages}.  Stage 1 is an outdoor map of size 100x100 pixels without any obstacles. Stage two is a mixed map of size 150x150 cells with static obstacles, which the agent knows. Stage 3 is an outdoor map of size 200x200 cells with known and additionally unknown static obstacles. Stage 4 is an indoor map of size 200x200 cells with known and unknown static obstacles. Stage 5 is an outdoor map of size 200x200 cells with known and unknown static obstacles and additionally unknown dynamic obstacles. Stage 6 is an indoor map of size 200x200 cells with known static obstacles and unknown static and dynamic obstacles. Stage 7 is almost the same as Stage 5 but with more unknown static and dynamic obstacles.
\\\\\noindent
The observations are processed by the DRL agent, which produces an action in the environment. Compared to our previous work \cite{kastner2021towards} training the DRL agent is not separated from the navigation system. Rather the full navigation stack is included inside the training loop. Although this might increase the overhead and extend the training due to more complex input, the agent should learn to synchronize better with the global planner and waypoint generator.
\\\\\noindent
While it is common to train only the local planner for local obstacle avoidance with DRL and integrate it as part of the full navigation stack, the proposed system already involves the global and waypoint generator and uses its input while in the training stage. Thereby, 6 different inputs where developed to test the extend to which these input will influence the behavior of the agent. The input of the different agents are listed in Tab. \cite{tab:agent-input}. 

\subsection{Agent and Neural Network Design}
\noindent In total, we propose six different agents, each with different observation spaces to study the effect of different inputs. The agent's input is listed in Tab. \ref{tab:agent-input}. The internal architecture and output layer are equal for all of them. They differ only in the input layer.

\begin{table}[htbp]

\centering
\setlength{\tabcolsep}{12pt}
\renewcommand{\arraystretch}{1}
\setlength{\tabcolsep}{.5\tabcolsep}
\footnotesize
\caption{Input of the different agents}
\begin{tabular}{@{}lccccc@{}}\toprule
Agent & Scan & Global Goal & Subgoal & Waypoints & Length  \\ 
\midrule
Agent 1&     0 - 359 & 360, 361 & 362, 363 & 364 - 463& -- \\
Agent 2 &    0 - 359 & 360, 361 & -- & -- & -- \\
Agent 3 &    0 - 359 & 360, 361 & 362, 363 & -- & 364 \\
Agent 4 &    0 - 359 & 360, 361 & 362, 363 & -- & -- \\
Agent 5 &    0 - 359 & 360, 361 & -- & -- & 362 \\
Agent 6 &    0 - 359 & 360, 361 & -- & 362 - 461 & -- \\
\bottomrule
\end{tabular}
\label{tab:agent-input}
\end{table}
\noindent All agents get as primary input the Lidar scan data and the global goal, represented as two values, the linear and angular distance from the odometry to the goal. All points are represented the same way as the global goal. Likewise, the optional subgoal is represented as a single point, whereas the global plan is represented in 2 different ways: as a representation of waypoints. From the global plan, every 5th point is extracted as a waypoint until 50 points are extracted in total. If the plan is not long enough to extract 50 points, the last extracted waypoint is used to fill up the waypoints list. Way 2 simplifies of the whole plan as a summed-up length of the plan. 
\\\\\noindent
The internal architecture is illustrated in Figure \ref{system}. For the body network, CNNs are used while the actor-critic network is designed using LSTM cells as \cite{memory_aided_drl_nav} showed the benefits of using memory-aided networks for navigation. The agent might be able to recognize movement directions and memorize older scan data for a better exploration of the area.\\
The output layer consists of 2 scalar values. Both are used to create a \textit{Twist} message for a 2D space. It consists of a linear velocity and an angular velocity. \\

\subsection{Reward Functions}
\noindent Since sparse rewards do not lead to fast convergence of the agent, we design our reward function to be dense and return a reward after each transition.
Negative rewards are only given for collisions or if the agent gets too close to a static or dynamic obstacle.
Positive rewards are given when the agent moves toward or reaches the target with a reasonable number of steps: the fewer steps required, the higher the reward. Equation 1 states the reward system of our agents. The reward function is the sum of all sub rewards
\begin{flalign}
    \label{eq:r_total}
    r^t = r^t_{gr} + r^t_{c} + r^t_{ga} + r^t_{sd} + r^t_{fgp} + r^t_{dgp} + r^t_{tc} + r^t_{adc}
\end{flalign}
\noindent Where  $r_{s}^t$ is the success reward for reaching the goal, $r_{c}^t$ is the punishment for a collision and both lead to episode ends.
 $r_{d}^t$ describes the reward for approaching the goal. Additionally, we introduce two safety rewards  $r_{ss}^t$ to help avoid static obstacles and $r_{sd}^t$ is meant for dynamic obstacles.
\begin{align}
    r^t_{gr} &= \Big\{ \begin{array}{ll} 45 & \text{, if goal reached}   \\ 0  &  \text{, otherwise}  \end{array} \\
    r^t_{c} &= \Big\{ \begin{array}{ll} -50 & \text{, if collided}   \\ 0  &  \text{, otherwise}  \end{array} \\
    r^t_{ga} &= \Big\{ \begin{array}{ll} 0.8 * diff^t_{robot, goal} & \text{, if $diff^t_{robot, goal} > 0$}\\ 0.6 * diff^t_{robot, goal} &  \text{, otherwise}  \end{array} \\
    r^t_{sd} &= \Big\{ \begin{array}{ll} -1.25 & \text{, if $\exists o \in O: d(p^t_{robot}, p^t_{o} < D_s)$}   \\ 0  &  \text{, otherwise}  \end{array}
\end{align}
\noindent
Reaching the goal gives a vast positive reward for the agent. This is the overall purpose of the agent. The reward for achieving this is set to 45. A collision results in a negative reward of -50. Getting closer to the goal seems good behavior, even though it is not like that in every case, such dead ends. That is why the reward should not be too high. Another is to keep a certain safe distance to obstacles. The agent should avoid driving just one mm away from obstacles. The calculation depends on $D_s$, the safe distance the agent should keep. It is set to 0.345m. The distance is calculated based on the center of the agent. As the agent has a radius of 0.3m, the safe distance between the agent surface to the obstacle surface is 4.5cm. A negative reward is given as soon as the agent is closer to an obstacle than the safe distance. Furthermore the rewards incorporating information about the global planner are defined as following:
\begin{align}
    r^t_{fgp} &= \Big\{ \begin{array}{ll} 0.1 * vel^t_{linear} & \text{, if $\min_{wp\in G} d(p^t_{wp}, p^t_{robot}) < 0.5m$}   \\ 0  &  \text{, otherwise}  \end{array} \\
    r^t_{dgp} &= \Big\{ \begin{array}{ll} 0.2 * diff^t_{robot, wp} & \text{, if $\frac{\min_{wp\in G} d(p^t_{wp}, p^t_{robot})}{diff^t_{robot, wp}} > 0$}   \\ 0  &  \text{, otherwise}  \end{array}\\
    r^t_{adc} &= -\frac{\left| vel^{t-1}_{angular} - vel^{t}_{angular}\right|^4}{1000}
\end{align}
with $diff^t_{x,y} = d(p^{t-1}_x, p^{t-1}_y) - d(p^{t}_x, p^{t}_y)$. The goal following reward $ r^t_{fgp}$ is weighted based on the linear velocity and is given if the agent is closer than 0.5m to the next point in the global plan. The distance to goal reward is another component of rewarding the agent for following the global plan is to reduce the distance to the closest point in the global plan.
Furthermore, agent navigation aims to drive smooth paths. That is why abrupt changes in the angular velocity are penalized with the reward $r^t_{adc}$.

\subsection{Training Hardware Setup}
\noindent Every agent is trained separately on one of two different systems. Table \ref{tab:specs} shows the hardware specifications of the systems. A docker image was created to perform the training on the systems. 

\begin{table}[htbp]
    \centering
    \begin{tabular}{lll}
        \toprule
        \textbf{Component} & \textbf{System 1} & \textbf{System 2} \\
        \midrule
        CPU & Ryzen Threadripper 1950X & Ryzen R7 2700X \\
        GPU & 2x NVIDIA RTX 2080TI & 1x NVIDIA RTX 2080TI \\
        RAM & 64GB & 40GB \\
        \bottomrule
    \end{tabular}
    \caption{Training System Specifications}
    \footnotesize This figure displays the specifications of the used computer systems for the training.
    \label{tab:specs}
\end{table}

\begin{figure*}[!h]
	\centering
	\includegraphics[width=\textwidth]{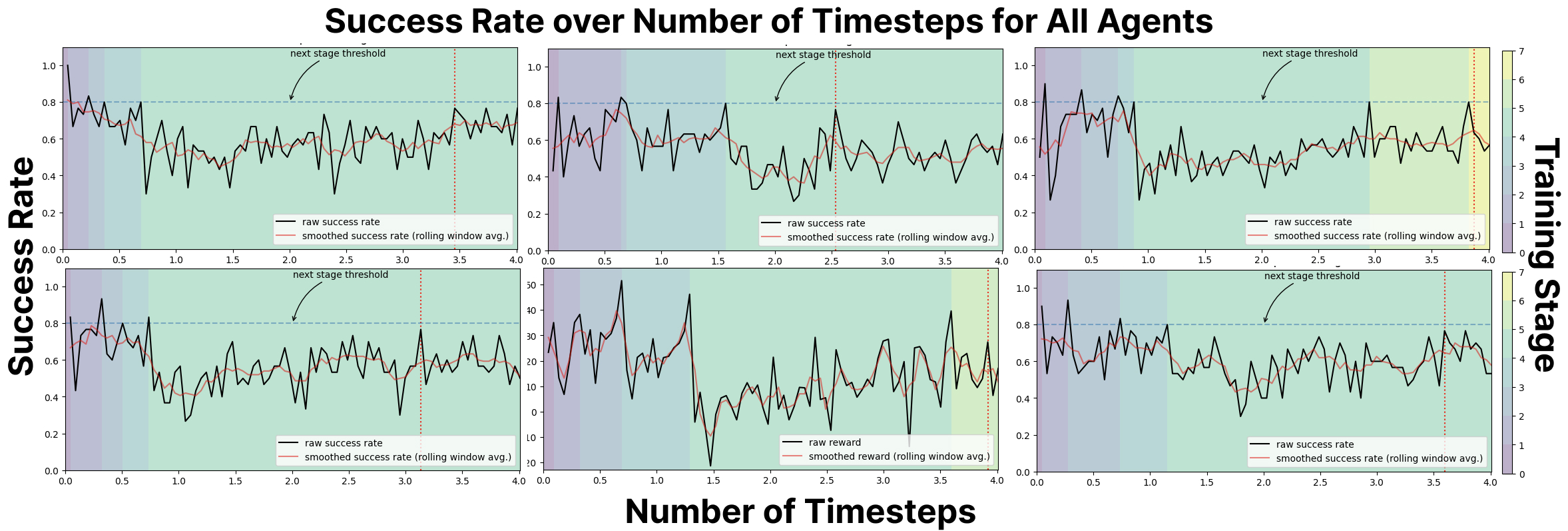}
	\caption{Imagination results for three different scenarios from four different models. The area covered by the blue mask is the imagination part. The results from four different models are trained on the different ground truth, namely 60x60, 60x60 Extended, 100x100, 100x100 Extended. The imaginations are based on thr observation and close to the real objects. The models trained with extended ground truth (denoted with 'Ext') predict the object more aggressively.}
	\label{fig:training}
\end{figure*}

\section{Evaluation}
\noindent In this chapter, we present the evaluation of our agents. The experiments are split into two categories. In the first part, we assess the training performance of all agents to assess the overhead and complexity of the training compared to a baseline agent with no additional input about the global planner and waypoint generator (Agent 2). In the second part, we compare our agents against baseline navigation approaches in terms of navigational metrics such as path length time to reach the goal etc. Therefore, we compared our agents against the classic local path planners Timed Elastic Bands (TEB) \cite{rosmann2015planning}, Dynamic Window Approach (DWA) \cite{khatib1986real}, and Model Predictive Control (MPC) \cite{rosmann2020online} as well as our proposed All-in-One Planner, which is able to switch between classic TEB and DRL planning \cite{kastner-aio}.

\subsection{Training Performance}
\noindent In order to evaluate the training process, the success rate of successfully completed tasks without collisions is investigated. The success rates are illustrated in Fig. \ref{fig:training}. Stage transitions are also indicated within that figure. Generally, the time points for reaching certain stages differ significantly among the agents. Not all agents reached the last stage 7. Only Agent 3 was capable of reaching that stage, rather late in the training. Agent 5 was able to reach stage 6 and all other agents only reached stage 5. Stage 5 was the first stage containing moving obstacles. Agent 1 and Agent 4 reached stage 5 earliest around training step 7M, and Agent 2 was latest around step 15.5M. The additional input seems to increase the learning speed in the lower stages. Namely, the agents with additional inputs can reach stage 5 much earlier than the baseline Agent 2 without additional input. Surprisingly, agents with similar input like Agent 1 and Agent 6 differ more than expected.
\\\\\noindent
As expected,the success rate of all agents include drops after reaching a new stage, indicating greater difficulty. Furthermore, the rate fluctuates for all agents, with some outlier drops in both directions. Although the agents have many similarities, some minor differences are observable. For example, Agent 2 seems to have a higher fluctuation than the other agents. 
Furthermore, the rate seems to stay at a level of around 60\% from step 25M onwards. In contrast, Agent 1 still has a slight improvement trend at the end of the training. Furthermore, the rate lies around 70\% in the last 5M steps. Agent 3 is the only agent that reached stage 7. However, some outlier runs might have caused those stage transitions. 
\\\\\noindent
In summary, Agent 1 has the highest level of success rate and reached stage 5 earliest and thus might be the best agent alongside Agent 3, which was able to reach the last stage 7. On the other hand, the baseline Agent 2 reached stage 5 the latest and has the lowest level of success rate. Furthermore, the agent seems not to improve anymore. These observations on the training metrics indicate a beneficial impact of additional input on training speed and also hint at a better performance.
\begin{figure*}[!h]
	\centering
	\includegraphics[width=0.99\textwidth]{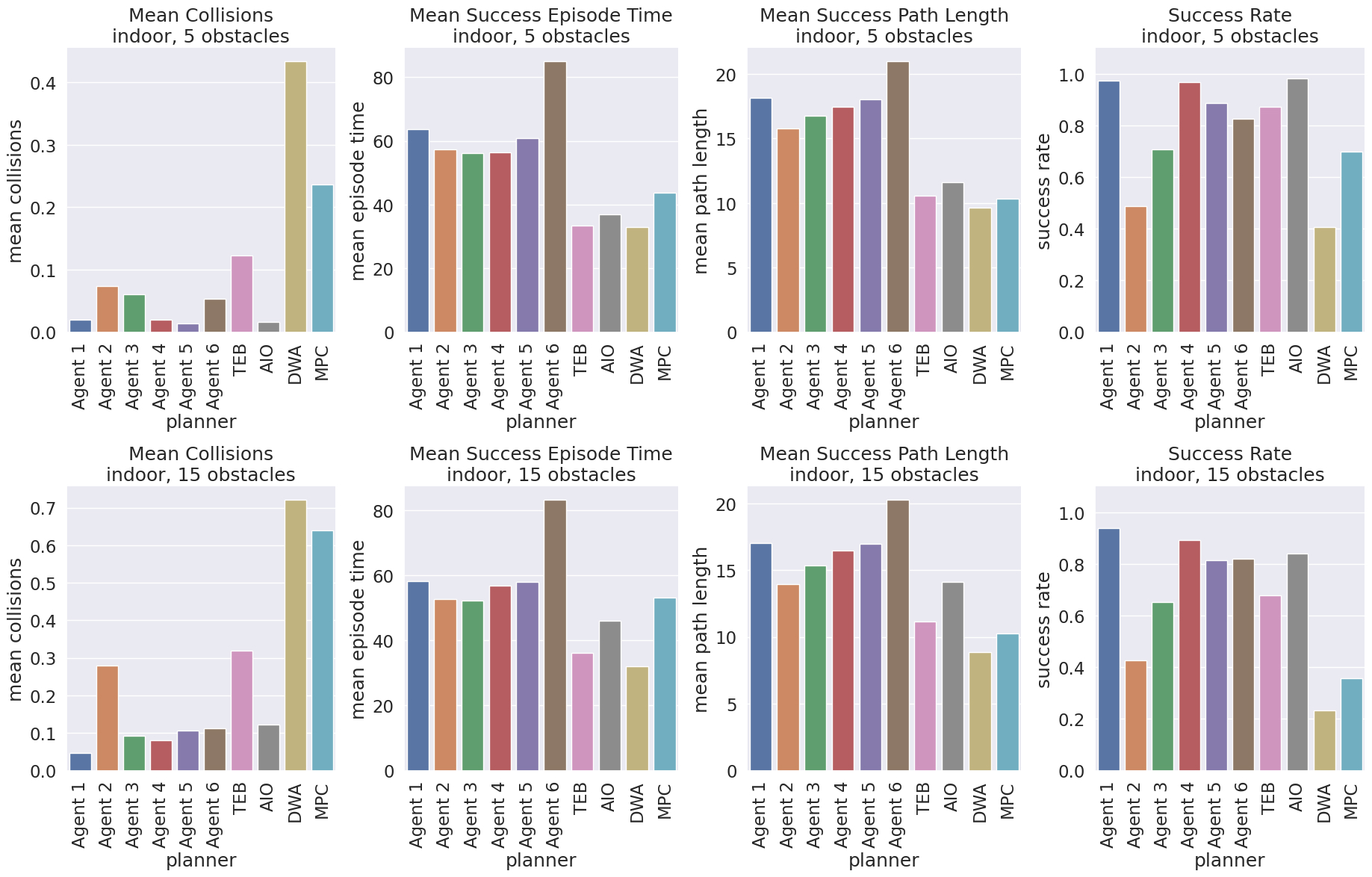}
	\caption{Imagination results for three different scenarios from four different models. The area covered by the blue mask is the imagination part. The results from four different models are trained on the different ground truth, namely 60x60, 60x60 Extended, 100x100, 100x100 Extended. The imaginations are based on thr observation and close to the real objects. The models trained with extended ground truth (denoted with 'Ext') predict the object more aggressively.}
	\label{fig:quanti}
\end{figure*}

\begin{figure}[!h ]
    \centering
    \includegraphics[width=0.99\linewidth]{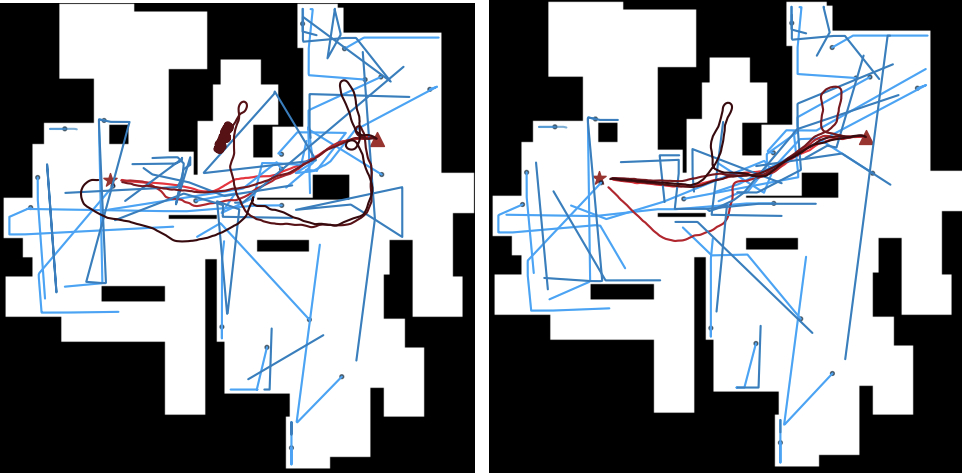}
    \caption{Our proposed agent can solve human-following and -guiding tasks within crowded environments. It was trained with additional semantic information like social states such as human talking, running etc. to reason and perform high level semantic tasks end-to-end.}
    \label{quali}
\end{figure}

\subsection{Navigational Performance}
\noindent After investigating the training metrics, the navigational performance of the agents are compared against baseline approaches. These include the model-based planners DWA \cite{khatib1986real}, TEB \cite{rosmann2015timed}, and MPC \cite{rosmann2019time}, as well as the AIO planner presented in our previous works, which is able to switch between TEB and DRL \cite{kastner-aio}. The evaluation is done by running 150 episodes for each agent in the same random scenarios and tasks. For each scenario, 150 episodes were performed. The comparison concentrates on success rates, path lengths, episode time, and collisions. For the qualitative evaluations of the navigational performance, we tested all approaches in three different scenarios: a) with 20 obstacles, b) with obstacle clusters, and c) with running obstacles with an obstacle velocity of up to 1m/s. The scenarios have a fixed start and goal position and the obstacles are moving according to the Pedsim social model \cite{helbing1995social}. The qualitative trajectories agents 3 and 5 are exemplarily illustrated in Fig. \ref{quali}. The timesteps are sampled every 100 ms and visualized within the trajectory of all approaches. The trajectories of the obstacles are marked with the start and end time in seconds. The episode ended once a collision occurred. Four metrics are considered for the base comparison: the success rate of reaching the goal without a collision, the mean number of collisions, the mean path length, and the mean time. An episode is considered unsuccessful when the agent collides with an obstacle, or a timeout happens. Figure \ref{fig:quanti} illustrates the results of all planners. 
\\\\\noindent
It is observed that Agent 1 performed best according to the success rate of 97.3\% and mean collisions of 0.02. whereas Agent 2 has the lowest success rate with 48.6\%. Surprisingly, Agent 3 performed rather severely with a success rate of just 70.6\% and mean collisions of 0.06. The other 4 agents performed similarly and can compete well with the AIO and TEB planner. All agents outperform the classic planners DWA and MPC except for the baseline Agent 2 and Agent 3. The path length of Agent 1 is slightly higher than that of Agent 2, 3 and 4, but in general, all path lengths are higher than those of the classic planners. 
\\\\\noindent
For scenarios with 15 obstacles the difference between the DRL-based agents and the classic baselines become even more noticeable. Whereas the success rates for the 6 agents did not change much, the rates of the baseline planners decreased more noticeably. The mean amount of collisions for Agent 2 has increased significantly from 0.94 to 5.18. The mean collisions of the other agents and planners have also increased, but not more than the drop in the success rate would imply. The path length, mean time and speed did not change in a substantially. 
\\\\\noindent
In general, Agent 1 performed best among the 6 agents, and an increase in obstacles did not significantly impact the performance. The performance is similar to the AIO planner on maps with 5 obstacles and slightly better on maps with 15 obstacles. As expected, Agent 2 performed worst of the 6 agents. Surprisingly, Agent 3 performed rather severely, although this agent was the only one which reached the last training stage. Especially Agent 2 and Agent 3 have comparably high path lengths. As they have the most timeouts, those agents sometimes might wander around the map without finding the goal. In some cases, the path lengths for scenarios with 15 obstacles are lower than for scenarios with 5 obstacles. 
\\\\\noindent
Fig. \ref{quali} exemplarily depicts the paths of agent 1 compared to the baseline agent 2. It is noticed that the agent with additional global information produces much more straightforward trajectories towards the goal, whereas the baseline agent with only Lidar information produce many roundabout paths. This indicates the better interoperation between global and local planner which also results in smoother trajectories. The navigation behavior of our planners is demonstrated visually in the supplementary video.

\section{Conclusion}
\noindent In this paper, we proposed a holistic DRL training pipeline incorporating all components and entities of the the ROS navigation stack typically used in industrial ground vehicles such as AGVs to improve synchronization between its entities. Rather than considering each entity of the navigation stack - global planner, waypoint generator, and local planner - separately, the training involves all entities and provides the DRL agent an enhanced understanding of them. Therefore, we integrated information about the global plan, the waypoint generator into the observation spaces of our trained agents. In total, we proposed six agents with different observation combinations to explore the effect of different input on the agents training and navigational performance. The additional information about the other entities could improve navigational performance and resulted in higher success rates, less collisions, and low path lengths compared to classic and learning-based baseline approaches. Future works aspire to explore the effect of additional input parameters such as semantic information about pedestrians or vehicles on the training and navigational performance. Furthermore, the approaches should be deployed towards real robots.


\addtolength{\textheight}{-1cm} 




\typeout{}
\bibliographystyle{IEEEtran}
\bibliography{main}

\end{document}